\PassOptionsToPackage{dvipsnames}{xcolor}
\documentclass[journal]{IEEEtran}
\IEEEoverridecommandlockouts

\usepackage{amsmath}
\usepackage{amssymb}
\usepackage{amsthm}
\usepackage{bm}
\usepackage{mathtools}
\usepackage{times}
\usepackage{bbm}

\usepackage{graphicx}
\usepackage{setup/support-caption}
\usepackage[font=footnotesize]{subcaption}
\usepackage{floatrow}
\usepackage{xspace}
\usepackage{blindtext}
\usepackage{algorithm} 
\usepackage{algorithmic}
\usepackage{tabularx}
\usepackage{booktabs}
\usepackage{orcidlink}
\usepackage{xcolor}

\usepackage{hyperref}
\hypersetup{
colorlinks=true,
linkcolor=blue,
filecolor=blue,      
urlcolor=blue,
citecolor=blue
}

\definecolor{michiganblue}{rgb}{0,0.153,0.310}
\definecolor{michiganmaize}{rgb}{1.0,0.796,0.020}
\definecolor{darkblue}{rgb}{0,0,0.7}

\DeclareMathOperator*{\argmin}{arg\,min}

\long\def\comment#1{}

\newcommand{\name}{ALPCAH }

\newcommand{\xmath}[1] {\ensuremath{#1}\xspace}

\newcommand{\blmath}[1] {\xmath{\bm{#1}}}
\newcommand{\paren}[1] {\xmath{\left(#1\right)}}

\newcommand{\normfrobr}[1] {\xmath{\| #1 \|_{\mathrm{F}}}} 
\newcommand{\algname}[1]{\textnormal{\textsc{#1}}}

\newcommand{\pbox}[1] {%
\makebox[0pt][r]{\raisebox{7mm}[0pt][0pt]{\small #1}}\ignorespaces}
\newcommand{\be} {\begin{equation}}
\newcommand{\ee}[1] {\label{#1}\end{equation}\pbox{#1}}
\newcommand{\eref}[1] {(\ref{#1})}

\newcommand{\green}[1]{\color{OliveGreen}{#1} \color{black}}
\newcommand{\fref}[1] {Fig.~\ref{#1}}
\newcommand{\aref}[1] {Alg.~\ref{#1}}

\newcommand{\sref}[1] {Sec.~\ref{#1}}
\newcommand{\tref}[1] {Thm.~\ref{#1}}
\newcommand{\tabref}[1] {Table~\ref{#1}}

\newcommand{\bs}{\begin{equation}\begin{split}}

\newcommand{\st}{\hspace{2mm} \text{s.t.} \hspace{2mm}}

\newcommand{\A}{\blmath{A}}
\newcommand{\C}{\blmath{C}}
\newcommand{\D}{\blmath{D}}

\newcommand{\E}{\blmath{E}}
\newcommand{\F}{\blmath{F}}
\newcommand{\G}{\blmath{G}}
\newcommand{\I}{\blmath{I}}
\renewcommand{\L}{\blmath{L}}
\newcommand{\R}{\blmath{R}}
\newcommand{\U}{\blmath{U}}
\newcommand{\V}{\blmath{V}}

\newcommand{\Y}{\blmath{Y}}
\newcommand{\Z}{\blmath{Z}}

\newcommand{\Lam}{\blmath{\Lambda}}
\newcommand{\Sig}{\blmath{\Sigma}}

\newcommand{\x}{\blmath{x}}

\newcommand{\w}{\blmath{w}}

\newcommand{\bmu}{\blmath{\mu}}
\newcommand{\bnu}{\blmath{\nu}}

\newcommand{\X}{\blmath{X}}
\newcommand{\y}{\blmath{y}}
\newcommand{\z}{\blmath{z}}
\newcommand{\e}{\blmath{e}}
\renewcommand{\r}{\blmath{r}}

\renewcommand{\u}{\blmath{u}}
\renewcommand{\v}{\blmath{v}}

\newcommand{\tr}{\text{Tr}}

\renewcommand{\xi}{\xmath{x_i}} 
\newcommand{\yi}{\xmath{\y_i}}

\newcommand{\hU} {\xmath{\hat{\U}}}
\newcommand{\hV} {\xmath{\hat{\V}}}
\newcommand{\hS} {\xmath{\hat{\Sig}}}
\newcommand{\hUd} {\xmath{\hat{\U}_{:,1:d}}}
\newcommand{\Ud} {\xmath{\U_{:,1:d}}}
\newcommand{\CX} {\xmath{\C_{\X}}}
\newcommand{\Cg} {\xmath{\C_{g}}}
\newcommand{\Cy} {\xmath{\C_{y}}} 
\newcommand{\Yg} {\xmath{\Y_{(g)}}}

\newcommand{\fhatd}{\xmath{f_{\hat{d}}}}
\newcommand{\Frac}[2]{#1/#2}

\let\originalPi=\Pi 
\renewcommand{\Pi}{\blmath{\originalPi}}

\newcommand{\0}{\blmath{0}} 
\newcommand{\bepsilon}{\boldsymbol{\epsilon}}

\newtheorem{definition}{Definition}
\newtheorem{theorem}{Theorem}

\usepackage{microtype}

\begin{document}

\title{
ALPCAH: Subspace Learning for \\ Sample-wise Heteroscedastic Data
}

\author{
\IEEEauthorblockN{ 
Javier Salazar Cavazos\orcidlink{0009-0009-1218-9836},~\IEEEmembership{Student Member,~IEEE},\\
Jeffrey A. Fessler\orcidlink{0000-0001-9998-3315},~\IEEEmembership{Fellow,~IEEE}, and 
Laura Balzano\orcidlink{0000-0003-2914-123X},~\IEEEmembership{Senior Member,~IEEE}
\\}
\IEEEauthorblockA{
ECE Department, University of Michigan, Ann Arbor, Michigan, United States\\
Email: \texttt{\{javiersc, fessler, girasole\}@umich.edu }}
}

\maketitle


\begin{abstract}
Principal component analysis (PCA) is a key tool in the field of data dimensionality reduction.
However, some applications
involve heterogeneous data that vary in quality
due to noise characteristics associated with each data sample.
Heteroscedastic methods aim to deal with such mixed data quality.
This paper develops a subspace learning method, named ALPCAH,
that can estimate the sample-wise noise variances
and use this information
to improve the estimate of the subspace basis
associated with the low-rank structure of the data.
Our method makes no distributional assumptions of the low-rank component
and does not assume that the noise variances are known.
Further, this method uses a soft rank constraint
that does not require subspace dimension to be known.
Additionally, this paper develops a matrix factorized version of ALPCAH,
named LR-ALPCAH, that is much faster and more memory efficient
at the cost of requiring subspace dimension to be known or estimated.
Simulations and real data experiments show the effectiveness
of accounting for data heteroscedasticity
compared to existing algorithms.
Code available at
\url{https://github.com/javiersc1/ALPCAH}.
\end{abstract}


\section{Introduction}
Many modern data-science problems require
learning an approximate signal subspace basis
for some collection of data.
This process is important for downstream tasks
involving the subspace basis coefficients such as 
classification \cite{classification}, 
regression \cite{regression}, 
and compression \cite{compression}.
More concretely,
lesion detection \cite{lesion_detection},
motion estimation \cite{motion_estimation},
dynamic MRI reconstruction \cite{dynamic_mri},
and image/video denoising \cite{video_denoising}
are practical applications involving the estimation of a subspace basis. 
In the modern ``big data'' world, a significant amount of data is collected
to solve problems, 
and this data tends to belong to a high-dimensional ambient space.
However, the underlying relationships between the data features
are often low dimensional
so the problem shifts towards finding low-dimensional structure in the data.

Some applications
involve heterogeneous data that vary in quality
due in part to noise characteristics associated with each data sample. 
Some examples of heteroscedastic datasets 
include environmental air data \cite{epa}, astronomical spectral data \cite{astronomical_data},
and biological sequencing data \cite{bacteria_data}.
In heteroscedastic settings,
the noisier data samples can significantly corrupt conventional basis estimates
\cite{asymptotic_pca}.
Subspace learning methods like probabilistic PCA (PPCA) \cite{ppca}
work well in the homoscedastic setting,
meaning when the data is the same quality throughout,
but fail to accurately estimate bases in the heteroscedastic setting
\cite{heppcat}.
This limitation is due to implicit assumptions
such as assuming that each sample's noise distribution is the same throughout (PPCA),
or in the case of the classical Robust PCA (RPCA) method \cite{RPCA},
that there are fewer outliers than good quality data samples.

\begin{figure}
  \centering
  \includegraphics[width=0.97\textwidth]{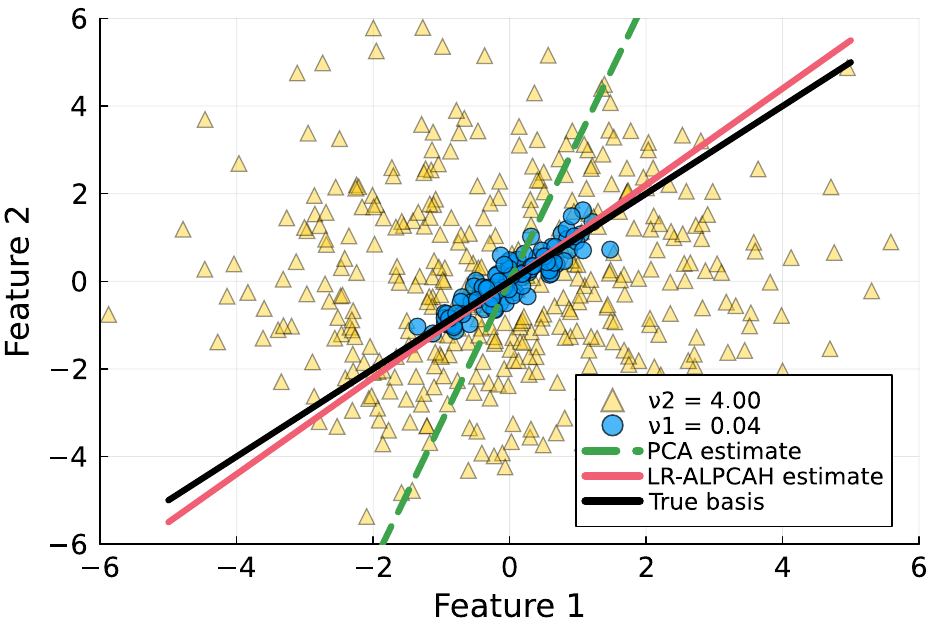}
  \caption{1D subspace with data consisting of two noise groups
  shown with circle and triangle markers.}
 \label{fig:subspace_example}
\end{figure}

A natural approach could be
to simply discard the noisiest samples to avoid this issue.
This approach 
requires the user to know the data quality,
which may be unavailable 
in practice.
That approach
also assumes that there is enough good data to estimate the basis,
but it is possible that a lack of good data 
requires using the noisy data,
especially if the subspace dimension is higher than the amount of good data.
Furthermore,
even the noisier samples can help improve the basis estimate
if properly modeled
\cite{heppcat},
so it is preferable to use all of the data available. 
This paper introduces subspace learning algorithms
that can estimate the sample-wise noise variances
and use this information in the model
to improve the estimate of the subspace basis
associated with the low-rank structure of the data.
See \fref{fig:subspace_example} for a visualization
where PCA fails to account for heteroscedasticity in a simple 2D data example,
but our LR-ALPCAH method more accurately finds the subspace basis.

The proposed subspace learning method, ALPCAH, first introduced in previous proceedings work \cite{alpcah}, 
allows for optional use of rank knowledge
via a low-rank promoting functional
and makes no distributional assumptions about the low-rank component of the data,
allowing it to achieve higher accuracy than current methods
without knowing the noise variances.
Moreover, we extend our previous proceedings work \cite{alpcah}
by developing an alternative formulation inspired by the matrix factorization literature
\cite{nonconvex_literature_overview},
that saves both memory and compute time
at the cost of requiring the subspace dimension to be known or estimated.

The paper is organized as follows.
Section \ref{paper:problem}
introduces the heteroscedastic problem formulation
for subspace learning and discusses related work.
Section \ref{paper:method_subspace}
introduces two subspace learning methods,
one with nuclear norm style low-rank regularization
originally introduced in our proceedings paper \cite{alpcah},
and an extension to a regularization-free maximum likelihood approach.
Section \ref{paper:experiments}
covers synthetic and real data experiments
that illustrate the effectiveness of these methods.
Finally, \sref{paper:conclusion}
discusses some limitations of our methods and possible extensions.
\section{Problem Formulation \& Related Works}
\label{paper:problem}

Let $\yi \in \mathbb{R}^{D}$ denote the data samples
for index $i \in \{1,\ldots,N \}$ given $N$ total samples,
and let $D$ denote the ambient dimension.
Let $\x_i$ represent the low-dimensional data sample
generated by $\x_i = \U \z_i$
where $\U \in \mathbb{R}^{D \times d}$ is an unknown subspace basis of dimension $d$
and $\z_i \in \mathbb{R}^{d}$ are the corresponding basis coordinates.
Collect the measurements into a matrix
$\Y = [\hspace{1mm} \y_1, \ldots, \y_N \hspace{1mm} ]$.
Then the heteroscedastic model we consider is
\begin{equation}
\yi = \x_i + \bepsilon_i
\quad \text{where} \quad
\bepsilon_i \sim \mathcal{N}(\0, \nu_i \I)
\label{eq:heteroscedastic_subspace_model}
\end{equation}
assuming Gaussian noise with variance $\nu_i$,
where \I denotes the $D \times D$ identity matrix.
We consider both the case where each data point may have its own noise variance,
and cases where there are
$G$ groups of data having shared noise variance terms
$\{ \nu_1,\ldots,\nu_G \}$.
\sref{paper:method_subspace}
proposes an optimization problem that estimates the heterogeneous noise variances
$\{\nu_i\}$
and the subspace basis \U.

\subsection{Heteroscedastic Impact on Subspace Quality}
Before describing the methods,
we illustrate
how heteroscedastic data
impacts the quality of the PCA subspace basis estimate
\hUd,
the first $d$ columns of \hU.
Let $\Y = \X + \E$
where $\X \in \mathbb{R}^{D \times N}$ is a rank-$d$ matrix
and $\E \in \mathbb{R}^{D \times N}$ is the noise matrix
where $ \E_{:,i} \sim \mathcal{N}(\0,\nu_i \I) \hspace{2mm} \forall j $.
Let $\Y = \hU \hS \hV'$ 
and $\X = \U \Sig \V'$
denote singular value decompositions of their respective matrices,
$\sigma_i(\A)$ denotes the $i$th singular value of \A.
Let $\| \A \|_2$ denote the spectral norm of matrix \A
and $\| \x \|_2$ denote the Euclidean norm of a vector \x.
The notation $a \lesssim b$ means $\exists k>0 \text{ s.t. } a \leq kb$.
By Wedin-Davis-Kahan $\sin\theta$ theorem \cite[p.~95]{probability},
it is known that
\begin{equation}
    \| \hUd \hUd' - \Ud \Ud' \|_2
    \leq  \frac{2\| \E \|_2}{\sigma_d(\X) - \sigma_{d+1}(\X)}.
    \label{eq:daviskahan}
\end{equation}

This inequality states that the maximum angle misalignment
between the latent subspace basis \Ud
and the SVD-estimated subspace \hUd
is bounded by the spectral norm of the noise matrix
over the spectral gap in matrix \X.
Assuming the elements in \E are zero mean and independent
(not necessarily identically distributed) random variables
it is known from \cite{normbound} that, in expectation, the spectral norm of $\E$ is bounded as
\begin{align}
    \mathbb{E}[\| \E \|_2] \lesssim &
    \max_i{\sqrt{\sum_j \mathbb{E}[E_{ij}^2]}}
    \nonumber \\
    & +
    \max_j{\sqrt{\sum_i \mathbb{E}[E_{ij}^2]}}
    + \sqrt[4]{\sum_{i,j}\mathbb{E}[E_{ij}^4]}.
    \label{eq:random_matrix_bound}
\end{align}
Because
$ \E_{:,i} \sim \mathcal{N}(\0, \nu_i \I) \hspace{2mm} \forall j $ in our application,
it can be verified that
\begin{align}
    \max_i{\sqrt{\sum_j \mathbb{E}[E_{ij}^2]}} &= \sqrt{ \nu^{(1)}_{\text{sum}}} \label{eq:rmt_bound1} \\
    \max_j{\sqrt{\sum_i \mathbb{E}[E_{ij}^2]}} &= \sqrt{D \nu_{\text{max}}} \label{eq:rmt_bound2} \\
    \sqrt[4]{\sum_{i,j}\mathbb{E}[E_{ij}^4]} &= \sqrt[4]{3D \nu^{(2)}_{\text{sum}}} \label{eq:rmt_bound3}
\end{align}
for
$\nu_{\text{max}} = \max_i \nu_i $
and $\nu_{\text{sum}}^{(k)} = \sum_i \nu_i^k$.
Let \CX correspond to the covariance matrix of \X,
i.e., $\CX = \frac{1}{N} \X \X'$.
Combining these bounds
with the property that $\sigma_{d+1}(\X) = 0$ for a rank-$d$ matrix
leads to the following result.
The subspace error, or more precisely,
the maximum angle separation between the
true subspace basis \Ud
and the estimated subspace basis \hUd
is bounded as follows
\begin{align}
    \mathbb{E}[\| & \hUd \hUd' - \Ud \Ud' \|_2^2] \lesssim \nonumber \\
    &\frac{
    \left( \sqrt{\nu^{(1)}_{\mathrm{sum}}} + \sqrt{D \nu_{\mathrm{max}}}
    + \sqrt[4]{3D \nu^{(2)}_{\mathrm{sum}}} \right) ^2
    }{N \sigma_d (C_X)}.
    \label{thm:pca_bound}
\end{align}
This upper bound indicates that the quality of the
subspace basis estimate \hUd
provided by the SVD of noisy data \Y, i.e., by conventional PCA,
could be degraded by heteroscedastic noise.
\fref{fig:bounds} in the appendix
provides empirical evidence for this claim. 
Thus, it can be advantageous to model the heteroscedasticity
and design a more robust PCA-like algorithm
that mitigates some of the effects of heteroscedastic noise
and achieves more accurate subspace basis estimates.

\subsection{Other heteroscedastic models}
\label{sec:other_models}

This paper focuses on heteroscedastic noise across the data samples.
There are other methods in the literature
that explore heteroscedasticity in different ways.
For example, HeteroPCA considers heteroscedasticity across the feature space
\cite{hetero_pca}.
One possible application of that model
is for data that consists of sensor information with multiple devices
that naturally have different levels of precision and signal to noise ratio (SNR).
Another heterogeneity model
considers the noise to be homoscedastic
and instead assume that the signal itself is heteroscedastic \cite{collas2021probabilistic}.
In that case, the power fluctuating signals are embedded in white Gaussian noise.
These different models each have their own families of applications.


\subsection{Probabilistic PCA (PPCA)}

PCA methods like PPCA \cite{ppca}
work well in the homoscedastic setting,
i.e., when the data is the same quality throughout,
but fail to accurately estimate the basis
when the data varies in quality,
e.g., in the heteroscedastic setting \cite{alpcah}.

Let $\C = \F \F' + \nu \I$ and observe that the model
\begin{align}
    \yi &= \F \z_i + \bepsilon \\
    \x_i &\sim \mathcal{N}(\0,\I), \hspace{1mm} \bepsilon \sim
    \mathcal{N}(\0,\nu \I), \hspace{1mm} \yi \sim \mathcal{N}(\0, \C) 
    \label{eq:ppca_model}
\end{align}
is similar to \eref{eq:heteroscedastic_subspace_model} in that
we have observation data \yi,
unobserved variables $\z_i$, factor matrix \F,
and noise term $\bepsilon$.
Then, for a covariance-type matrix
$\Cy = \sum_i \yi \yi'$ formed from data samples \Y,
the negative log-likelihood is
\begin{equation}
\mathcal{L}(\F,\nu) = -\frac{N}{2}(d\log(2\pi)+\log(|\C|)+\tr(\C^{-1} \Cy)),
\label{eq:ppca_cost}
\end{equation}
where $|\cdot|$ and $\tr(\cdot)$ denote matrix determinant and trace, respectively.
After estimating \F and $\nu$ by minimizing \eref{eq:ppca_cost},
PPCA finds the subspace basis
by orthogonalizing \F.
Because
$\bepsilon \sim \mathcal{N}(\0,\nu \I)$
is identically distributed across all data samples,
PPCA does not account for heterogeneous data samples.

\subsection{Robust PCA (RPCA)}

Robust PCA (RPCA) \cite{RPCA} decomposes the data matrix
$\Y = \X + \E$
into a low-rank component \X and an outlier matrix \E
by the following optimization problem:
\begin{equation}
\argmin_{\X,\E} 
\paren{ \lambda \Vert \X \Vert_* + \Vert \E \Vert_{1,1} }
\st \Y = \X + \E
\end{equation}
where
$\| \X \|_* = \sum_{i=1}^{\hidewidth \min(M,N) \hidewidth} \sigma_i(\X)$
and $\|\E\|_{1,1} = \sum_{i,j} |E_{ij}|$.
RPCA finds the subspace basis by iteratively applying an SVD to $\X$ to soft threshold the singular values.
Here, the term $\|\E\|_{1,1}$ encourages sparsity
and so captures noise in the data matrix
by assuming there is a sparse collection of outliers.
This modeling assumption may not be true in some applications;
for instance, low-quality and abundant commercial sensors
are often combined with fewer high-quality sensors.
Ref.~\cite{alpcah}
illustrated the limitations of RPCA
in the heteroscedastic regime.

\subsection{Weighted PCA (WPCA)}

Given data samples $\{ \y_1, \ldots, \y_N\}$ and weights $\{w_1, \ldots, w_N\}$,
the weighted PCA (WPCA) approach \cite{wpca}
for modeling heteroscedastic data
forms the following weighted sample covariance matrix

\begin{equation}
\Cy(\w)= \sum_{i=1}^N w_i (\yi \y^T_i),
\end{equation}
where a natural choice for the weights is $w_i = \nu_i^{-1}$.
WPCA finds the subspace basis
by orthogonalizing $\Cy(\w)$ via eigenvalue decomposition (EVD).
However, the noise variances may not be known,
e.g., unknown origin of the dataset or unavailable data sheet for physical sensors.

\subsection{Heteroscedastic PPCA Technique (HePPCAT)}

To our knowledge, besides ALPCAH,
there is only one sample-based heteroscedastic PCA algorithm
that estimates unknown noise variances.
The Heteroscedastic Probabilistic PCA Technique (HePPCAT) \cite{heppcat}
builds on the PPCA formulation.
For $n_1 + \ldots + n_G = N$ data samples from $G$ noise groups,
the model is described as

\begin{equation}
\y_{g,i} = \F \z_{g,i} + \bepsilon_{g,i}
\quad i \in \{ 1,\ldots,n_G\}, \ g \in \{1,\ldots,G\}
\label{eq:heppcat_model}
\end{equation}
for factor scores $\z_{g,i} \sim \mathcal{N}(\0,\I)$,
noise terms $\bepsilon_{g,i} \sim \mathcal{N}(\0,v_g \I)$, 
and points $\y_{g,i} \sim \mathcal{N}(\0, \Cg)$
where
$\Cg = \F \F' + v_g \I$ for factor matrix \F.
Then, the negative log-likelihood model to optimize is the following
\begin{equation}
    \mathcal{L}(\F,\v) = 
    \frac{1}{2} \sum_{g=1}^G [n_g \ln \det (\Cg)^{-1} - \tr \{\Yg^T (\Cg)^{-1} \Yg\}]
    \label{eq:heppcat_cost}
\end{equation}
where \Yg denotes the submatrix of \Y
that consists only of data samples belonging to the $g$th noise group,
and $ \v = (v_1, \ldots, v_G) $
denotes the unknown nose variances for each group.
Being a factor analysis method,
HePPCAT makes Gaussian assumptions about the basis coefficients $\z_{l,i}$
that may not be a good model for some datasets.
Additionally, HePPCAT requires the rank parameter $d$ associated with
the latent signal matrix \X
to be estimated or known a priori.
\section{Proposed Subspace Learning Methods} 
\label{paper:method_subspace}

This section introduces the ALPCAH formulation for subspace learning.
Since nuclear norm computation is expensive for big data applications due to SVD computations,
we take inspiration from the matrix factorization literature
and additionally develop LR-ALPCAH to be a fast and memory efficient 
alternative to ALPCAH.

\subsection{ALPCAH}
\label{sec:alpcah}
For the measurement model
$\yi \sim \mathcal{N}(\x_i, \nu_i \I)$
in \eqref{eq:heteroscedastic_subspace_model},
the probability density function for a single data vector \yi is
\begin{equation}
 \frac{1}{\sqrt{(2\pi)^D |\nu_i \I |}}
 \exp{ [-\frac{1}{2} (\yi-\x_i)' (\nu_i \I)^{-1} (\yi-\x_i)] }.
\end{equation}
For uncorrelated samples, after dropping constants, 
the joint log likelihood of all data $\{\yi\}_{i=1}^{N}$ is the following
\begin{equation}
    \sum_{i=1}^{N} -\frac{1}{2} \log |\nu_i \I|
    -\frac{1}{2} (\yi - \x_i)' (\nu_i \I)^{-1} (\yi - \x_i).
\end{equation}
Let
$\Pi = \mathrm{diag}(\nu_1,\ldots,\nu_N) \in \mathbb{R}^{N \times N}$
be a diagonal matrix representing the (typically unknown) noise variances.
Then, the negative log likelihood in matrix form is
\begin{align}
&
\frac{D}{2} \log |\Pi| + \frac{1}{2} \text{Tr}[(\Y - \X) \Pi^{-1} (\Y - \X)']
\nonumber\\=&
\frac{D}{2} \log |\Pi| +
\frac{1}{2} \normfrobr{ (\Y - \X) \Pi^{-1/2} }^2,
\label{eq:ML}
\end{align}
using trace lemmas.
When both $\Pi$ and \X are unknown,
pursuing maximum-likelihood estimation
with
\eref{eq:ML}
would lead to degenerate solutions.
Thus, regularization is necessary to promote a low-rank solution.
In this work, we use a functional modified from the nuclear norm regularizer
to encourage
the estimate of \X to be low rank.

The optimization problem
used by ALPCAH
for the heteroscedastic model is
\begin{equation}
    \argmin_{\X,\Pi} \lambda f_{\hat{d}}(\X)
    + 
    \frac{1}{2} \normfrobr{ (\Y - \X) \Pi^{-1/2} }^2
    + \frac{D}{2} \log |\Pi|
    \label{eq:alpcah_cost_function},
\end{equation}
where $f_{\hat{d}}(\X)$ is a novel functional \cite{pssv}
that promotes low-rank structure in \X,
$\hat{d}$ is the rank parameter,
and
$\lambda \in \mathbb{R}^{+}$ is a regularization parameter.
In the following, we introduce our algorithm called \name
(\textbf{A}lgorithm for \textbf{L}ow-rank regularized \textbf{PCA} for \textbf{H}eteroscedastic data)
for solving \eqref{eq:alpcah_cost_function}. 
Since $\X$ represents the denoised data matrix,
the subspace basis is calculated by SVD on the optimal solution
from \eqref{eq:alpcah_cost_function} and extracting the first $\hat{d}$ left singular vectors
so that
$\Hat{\X} = \sum_i \Hat{\sigma}_i \Hat{\u}_i \Hat{\v}_i'$
and thus
$\hU = [\Hat{\u}_1,\ldots,\Hat{\u}_{\hat{d}}]$.
The low-rank promoting functional we use
is the sum of the tail singular values
defined as 
\begin{equation}
\fhatd(\X) \triangleq \sum_{i=\hat{d}+1}^{\text{min}(D,N)} \sigma_i (\X)
= \| \X \|_* - \| \X \|_{\mathrm{Ky-Fan}(\hat{d})}
\end{equation}
where
$\| \cdot \|_*$ denotes the nuclear norm,
and $\| \cdot \|_{\mathrm{Ky-Fan}(\hat{d})}$ denotes the Ky-Fan norm \cite{norms}
defined as the sum of the first $\hat{d}$ singular values.
For $\hat{d}=0$, $f_0(\X) = \| \X \|_*$.
For a general $\hat{d}>0$,
$\fhatd(\X)$ is a nonconvex difference of convex functions.
We use the functional \fhatd
instead of the nuclear norm
since we empirically found that the nuclear norm tends to over shrink the singular values of \X
in the heteroscedastic setting.
Here, the rank parameter $\hat{d} \ll D$ is either known beforehand,
estimated using methods like row permutations \cite{permutationMethods}
or sign flips \cite{signFlips},
or intentionally over-parameterized.

\begin{definition}
Let $\A \in \mathbb{R}^{D \times N}$ be a rank $k$ matrix
such that its decomposition is $\mathrm{SVD}(\A) = \U_A \D_A \V_A '$
where $\D_A = \mathrm{diag}(\sigma_1 (\A), \ldots, \sigma_{\min(D,N)} (\A))$.
Let the soft thresholding operation be defined as
$\mathcal{S}_{\tau}[x] = \mathrm{sign}(x) \max(|x| - \tau, 0)$
for some threshold $\tau > 0$.
Decompose $\D_A$ such that
$\D_A = \D_{A1} + \D_{A2}
= \mathrm{diag}(\sigma_1 (\A),\ldots,\sigma_{\hat{d}}(\A),0,\ldots,0)
+ \mathrm{diag}(0,\ldots,0,\sigma_{\hat{d}+1}(\A),\ldots,\sigma_{N} (\A))$.
Then, the proximal map for \fhatd
is the tail singular value thresholding operation
\cite{pssv}:
\begin{equation}
\mathrm{TSVT}(\A, \tau, \hat{d}) \triangleq \U_A \, (\D_{A1} + \mathcal{S}_{\tau} [\D_{A2}] ) \, \V_A'.
\label{eq:tsvt}
\end{equation}
\end{definition}

Although the proximal operator for \fhatd is provided
in \eqref{eq:tsvt},
it is unclear how one would apply a proximal gradient method (PGM)
directly to \eqref{eq:alpcah_cost_function}
due to the product of \X and \Pi.
One could apply a block coordinate decent approach
that alternates between updating \X
using a PGM,
and updates the diagonal elements of \Pi
using a closed-form solution.
The PGM update of \X
could cause slow convergence
because the Lipschitz constant
of the gradient of the smooth term
$g(\X)$
is the reciprocal
of the smallest diagonal element of \Pi,
which could be quite large,
leading to small step sizes.
Thus, instead we optimize
\eref{eq:alpcah_cost_function} 
using the inexact augmented Lagrangian method known
as the alternating direction method of multipliers (ADMM) \cite{admm}
that introduces auxiliary variables
to convert a complicated optimization problem
into a sequence of simpler optimization problems.

Defining the auxiliary variable
$\Z = \Y - \X$,
the augmented penalty parameter
$\mu \in \mathbb{R}$,
and dual variable
$\Lam \in \mathbb{R}^{D \times N}$,
the augmented Lagrangian, as defined in \cite{alm_theory}, is
\begin{align}
\mathcal{L}_{\mu}(\X, & \Z, \Lam, \Pi) =
\lambda \fhatd(\X) + \frac{1}{2} \normfrobr{ \Z \Pi^{-1/2} } ^2
+ \frac{D}{2} \log |\Pi |
\nonumber \\ 
&+ \langle \Lam, \Y - \X - \Z \rangle + \frac{\mu}{2} \normfrobr{ \Y - \X - \Z }^2
,
\label{eq:auglagrangian}
\end{align}
where $\langle \cdot, \cdot \rangle$
denotes the Frobenius inner product
between two matrices.


Performing a block Gauss-Seidel pass for each variable
in \eref{eq:auglagrangian}
results in the following closed-form updates
\begin{align}
\Z_{t+1} &= \argmin_{\Z} \mathcal{L}_{\mu}(\X_t, \Z, \Lam_t, \Pi_t)
\nonumber \\
&= [\mu \, (\Y - \X_t) + \Lam_t] \, (\Pi_t^{-1} + \mu \I )^{-1}
\label{eq:zupdate} \\
\X_{t+1} &= \argmin_{\X} \mathcal{L}_{\mu}(\X, \Z_t, \Lam_t, \Pi_t)
\nonumber \\
&= \mathrm{TSVT}(\Y - \Z_t + \frac{1}{\mu} \Lam_t, \frac{\lambda}{\mu}, \hat{d})
\label{eq:xupdate}\\
\Lam_{t+1} &= \Lam_t + \mu \, (\Y - \X_t - \Z_t)
\label{eq:lambdaupdate}
\end{align}
for current iteration pass $t$. Each pass is run for $T$ total iterations.
When we treat data sample \yi as having its own unknown noise variance,
then the variance update is
\begin{align}
    \Pi_{t+1} &= \argmin_{\Pi} \mathcal{L}_{\mu}(\X_t, \Z_t, \Lam_t, \Pi)
    = \frac{1}{D} \Z_t' \Z_t \odot \I.
    \label{eq:var_update}
\end{align}
For the case when the data points have grouped noise variances, let
$g \in \{1,\ldots, G\}$ signify the $g$th noise group out of $G$ total groups with $n_g$ denoting the number of samples in the $g$th group;
then the grouped noise variance update instead becomes
\begin{equation}
\nu_g = \frac{1}{D n_g} \normfrobr{ \Z_{(g)} } ^2
= \frac{1}{D n_g} \normfrobr{ \Y_{(g)} - \X_{(g)} } ^2
\label{eq:var_update2}
\end{equation}
where the notation \Yg denotes the submatrix of \Y
that consists only of data samples belonging to the $g$th noise group.

\subsubsection{Convergence with known variance}
Consider the cost function for the case when the variances $\Pi$ are known.
The formulation consists of a two-block setup written as 
\begin{equation}
\argmin_{\X, \Z}  \lambda \fhatd(\X)
+ \frac{1}{2} \normfrobr{ \Z \Pi^{-1/2} } ^2  \st \Y = \X + \Z.
\end{equation}
\begin{theorem}
\label{thm:alpcah}
Let $\Psi(\X,\Z) = f(\X) + g(\Z)$
where $f(\X) = \lambda \fhatd(\X)$
and $g(\Z) = \frac{1}{2} \normfrobr{ \Z \Pi^{-1/2} } ^2 $.
Let $\nu_i \geq \epsilon > 0 \hspace{3mm} \forall i$.
Assuming that $\mu$ in \eqref{eq:auglagrangian}
satisfies $\mu > 2 L_g = 2 \|\Pi^{-1}\|_2 $, the sequence 
$\{(\X_t,\Z_t,\Lam_t,\Pi)\}_{i=1}^T$ generated by ADMM
in \eref{eq:zupdate} \eref{eq:xupdate} \eref{eq:lambdaupdate} \eref{eq:var_update}
converges to a KKT (Karush–Kuhn–Tucker) point
of the augmented Lagrangian
$\mathcal{L}_{\mu}(\X,\Z,\Lam, \Pi)$ with fixed $\Pi$.
\end{theorem}

\begin{proof}
ADMM convergence for nonconvex problems has been explored for two-block setups \cite{admm_nonconvex}.
The functional $f(\X)$ is a proper, lower semi-continuous function
since it is a sum of continuous functions.
The function $g(\Z)$ is a continuous differentiable function
whose gradient is Lipschitz continuous with modulus of continuity
$L_g = \|\Pi^{-1}\|_2$ .
By definition $g(\Z) = \nu_1^{-1/2} Z_{11} + \nu_1^{-1/2} Z_{21} + \ldots + \nu_{N}^{-1/2} Z_{DN}$.
Since $g(\Z)$ is a polynomial equation,
its graph is a semi-algebraic set.

Let
$\G = \X'\X \in \mathbb{R}^{N \times N}$.
Then, by Cayley Hamilton theorem,
the characteristic polynomial of \G is
$p_{G} (z) = z^N + c_{n-1}(\G) z^{N-1} + \ldots + c_1 (\G) z + c_0$
for constants $c_i \in \mathbb{R}$ and polynomial degree $N$.
Let $\lambda$ denote an eigenvalue of \G
which implies $p_G (\lambda) = 0$.
Then, the set
$\mathcal{S}_G = \{ \lambda \hspace{1mm} | \hspace{1mm} p_G (\lambda) = 0 \}$
is semi-algebraic since it is defined by polynomial equations.
Note that $\lambda_i(\G) = \sigma_i^2(\X)$
since \G is the Gram matrix of \X.
The set
$\mathcal{S}_X = \{ \sigma \hspace{1mm} | \hspace{1mm} \sigma^2
= \lambda \in S_G, \sigma \geq 0 \} = \{ \sigma_1, \ldots, \sigma_N \}$
is semi-algebraic as it is expressed in terms of polynomial inequalities.
Expressing $h(\X) = \|\X\|_* = h(\sigma_1, \ldots, \sigma_N) $,
its graph $h = \{ (\mathbf{\sigma}, f(\mathbf{\sigma})) \} $ is semi-algebraic
and thus by extension so is the nuclear norm. 

By Tarksi-Seidenburg theorem \cite[p.~345]{tarski-seidenburg},
defining the map
$\Phi : \mathbb{R}^n \rightarrow \mathbb{R}^{\hat{d}}$
that retains the first $\hat{d}$ singular values of $\mathcal{S}_X$,
the set $\Phi (\mathcal{S}_X) = \{ \sigma_1, \ldots, \sigma_{\hat{d}} \}$
is semi-algebraic and thus so is
$q(\X) = \|\X\|_{\mathrm{Ky-Fan}(\hat{d})}$.
A finite weighted sum of semi-algebraic functions
is known to be semi-algebraic \cite{attouch_presentation}
and so $f(\X) = h(\X) - q(\X)$ is semi-algebraic. 
Since the functions $f(\X)$ and $g(\Z)$ are lower, semi-continuous
and definable on an o-minimal structure such as semi-algebraic \cite{o-minimal},
it follows that $\Psi(\X,\Z) = f(\X) + g(\Z)$
is a Kurdyka-Lojasiewicz function \cite{attouch_presentation}.
Thus the sequence $\{ (\X_t, \Z_t, \Lam_t, \Pi) \}_{i \in \mathbb{N}}$
converges to a KKT point by
\cite[Thm.~3.1]{admm_nonconvex}.
\end{proof}

\subsection{LR-ALPCAH}
\label{sec:lr-alpcah}
The main compute expense for the ALPCAH algorithm
is the SVD operations used in every iteration of complexity $\mathcal{O}(DN\min{(D,N)})$.
To reduce computation,
we take inspiration from the matrix factorization literature
\cite{matrix_factorization}
and factorize $\X \in \mathbb{R}^{D \times N} \approx \L \R'$
where $\L \in \mathbb{R}^{D \times \hat{d}}$
and $\R \in \mathbb{R}^{N \times \hat{d}}$
for some rank estimate $\hat{d}$. Using the factorized form,
we propose to estimate \X
by solving for \L and \R in the following optimization problem
\begin{align}
    \hat{\L}, \hat{\R}, \hat{\Pi} = & \argmin_{\L,\R,\Pi} \ f(\L,\R,\Pi)
    \nonumber \\
    f(\L,\R,\Pi) =
    &\frac{1}{2} \normfrobr{ (\Y - \L \R') \Pi^{-1/2} }^2 + \frac{D}{2} \log |\Pi|.
    \label{eq:lr_alpcah}
\end{align}

This version is a maximum-likelihood estimator
of $\Pi$ and the factors \L and \R.
This comes from a modified model \eref{eq:heteroscedastic_subspace_model} where
\begin{equation}
    \y_i = \L \r_i + \bepsilon_i , \quad \bepsilon_i \sim N(\0, \nu_i \I),
\end{equation}
where $\r_i$ denotes the $i$th column of \R.
We call this version LR-ALPCAH
given the prevalence of $\L\R'$ notation in the matrix factorization literature.
The crucial difference between ALPCAH and LR-ALPCAH
is that ALPCAH uses a ``soft'' low rank constraint
through the regularization penalty $\lambda$ with optional usage of $\hat{d}$,
whereas LR-ALPCAH uses a ``hard'' low rank constraint
since $\L$ and $\R$ rigidly contain $\hat{d}$ columns. 

We solve this optimization problem
using alternating minimization \cite{altmin}
to solve each sub-block, resulting in the following updates:

\begin{align}
    \L_{t+1} &= \argmin_{\L} f(\L, \R_t, \Pi_t)
    \nonumber \\
    &= \Y \Pi_t^{-1} \R_t (\R_t' \Pi_t^{-1} \R_t)^{-1}
    \label{eq:lr_l_update}\\
    \R_{t+1} &= \argmin_{\R} f(\L_t, \R, \Pi_t)
    \nonumber \\
    &= \Y' \L_t (\L_t' \L_t)^{-1} \label{eq:lr_r_update} \\
    \Pi_{t+1} &= \argmin_{\Pi} f(\L_t, \R_t, \Pi)
    \implies \nonumber \\
    \e_j' \Pi_{t+1} \e_j &= D^{-1} \| (\Y - \L_t \R_t') \e_j\|_2^2, \ \forall j,
    \label{eq:lr_pi_update}
\end{align}
where $\e_j$ denotes the $j$th standard canonical basis vector
that we use to select the $j$th column of some matrix.
The $\Pi_t$ update \eref{eq:lr_pi_update} is the same as  \eref{eq:var_update}
in that each point is treated as having its own noise variance
and both equations perform the same operation.
This implementation requires less computation and memory
since the matrix $\Z_t' \Z_t$ is not formed.
One can substitute \eref{eq:lr_pi_update} 
with \eref{eq:var_update2} if noise grouping is known.

Since this is a nonconvex problem,
initialization will play a key role in the success of optimization.
First, we initialize the $\L_t$ and $\R_t$ matrices
with the following spectral approach:
\begin{align}
\text{Spectral Init}(\Y)
= \hU \hS \hV' &\approx
(\hU_{1:\hat{d}} \hS_{1:\hat{d}}^{1/2}) (\hS_{1:\hat{d}}^{1/2} \hV_{1:\hat{d}}')
\nonumber \\
&= (\L_0) (\R_0').
\label{eq:spectral_init}
\end{align}
This initialization
from 
the matrix factorization literature
\cite{nonconvex_literature_overview} \cite{procrustes_flow}
is a natural approach
due to the Eckart-Young theorem \cite{eckartyoung}
that shows $\L_0$ and $\R_0$ are the best rank-constrained matrices that solve
\begin{equation}
    \argmin_{\L,\R} \normfrobr{ \Y - \L \R' }
    \text{ subject to } \text{rank}(\L),\text{rank}(\R) \leq \hat{d}.
    \label{eq:eckart_young}
\end{equation}
Second,
we initialize the noise variances
using the Euclidean norms of the columns of the residual
$\Y - \L_0 \R_0'$ with \eref{eq:lr_pi_update}.
Finally,
we apply alternating minimization
to update $\L_t$ and $\R_t$ matrices at current iteration $t$
via \eref{eq:lr_l_update} \eref{eq:lr_r_update} \eref{eq:lr_pi_update}.

Since $\L_t$ is not semi-unitary
but 
has the same range as
\U,
we apply Gram-Schmidt orthnormalization to
the final $\L_t$ matrix to estimate the subspace basis,
as described
in \aref{alg:lr-alpcah}.
The matrix inversions used in the $\L_t$ and $\R_t$ updates
involve $d \times d$ matrices
that are relatively small and thus computationally feasible for many practical problems
given complexity $\mathcal{O}(\hat{d}^3)$ knowing that $\hat{d} \ll \min{(D,N)}$. Combining the matrix multiplications and inversions, LR-ALPCAH has a per-iteration complexity of $\mathcal{O}(DN\hat{d} + \hat{d}^3)$. This is in contrast to ALPCAH with per-iteration complexity $\mathcal{O}(DN \min{(D,N)})$ due to the SVD computations.

\begin{algorithm} 
    \renewcommand{\thealgorithm}{LR-ALPCAH}
    \caption{ (\href{https://github.com/javiersc1/ALPCAH}{github.com/javiersc1/ALPCAH}) \\
    (unknown variances, unknown quality noise grouping)}
    \label{alg:lr-alpcah}
    \begin{algorithmic}
        \STATE \textbf{Input:} $\Y \in \mathbb{R}^{D \times N}$: data, $\hat{d} \in \mathbb{N}^{*}$: rank estimate
        \STATE \textbf{Opt:} $T\in \mathbb{N}^{*}$: iterations, $\epsilon \in \mathbb{R}^{+}$: variance noise floor
        \STATE \textbf{Output:} $\U \in \mathbb{R}^{D \times \hat{d}}$: subspace basis,
        $\X \in \mathbb{R}^{D \times N}$: low-rank estimated data,
        $\bnu \in \mathbb{R}^{+}$: estimated noise variances
            \\ { \green{ // sample mean to de-mean data }}
                \STATE $\bmu \gets \frac{1}{N} \Y \bm{1} $ \hfill 
            \\ { \green{// the method assumes linear subspaces only } }
                \STATE $\Y \gets \Y -  \bm{1}' \bmu$ \hfill 
            \\ { \green{ // initialize matrices by  \eref{eq:spectral_init} } } 
              \STATE $\L_0, \R_0 \gets \algname{SpectralInit}(\Y, \hat{d})$  \hfill
              \\ { \green{ // compute noise variances from residuals $\Y - \L_0 \R_0'$ }}
              \\ { \green{ // $\e_j$ is canonical basis vector }} 
             \STATE $\nu_0 \gets \max_{j=1,\ldots,N} \frac{1}{D} \| (\Y - \L_0 \R'_0) \e_j \|_2^2$
             \STATE $\nu_0 \gets \max(\nu_0, \epsilon)$
             \STATE $\Pi^{-1}_{0} \gets (1/\nu_0) \I$
             \\ { \green{ // update $\L,\R,\Pi$ matrices using \eref{eq:lr_l_update} \eref{eq:lr_r_update} \eref{eq:lr_pi_update} } }
        \FOR{$t = 1,\dots,T$}
             \STATE $\L_t \gets \Y \Pi^{-1}_{t-1} \R_{t-1} (\R'_{t-1} \Pi^{-1}_{t-1} \R_{t-1})^{-1}$ 
             \STATE $\R_t \gets \Y_{t-1}' \L_{t-1} (\L'_{t-1} \L_{t-1})^{-1}$
             \STATE $\nu_j \gets \max(\frac{1}{D} \|  (\Y - \L_{t-1} \R'_{t-1}) \e_j \|_2^2, \, \epsilon)
             ,\ j = 1,\ldots,N$
             \STATE $\Pi^{-1}_{t} \gets \text{Diagonal}(1/\nu_j)$
        \ENDFOR
        \\  \green{ // form subspace basis from final left factor} 
        \STATE $\U \gets \algname{GramSchmidt}(\L_T)$
        \\ { \green{ // construct de-meaned low-rank estimate } }
        \STATE $\X \gets \L_T \R_T'$ \hfill 
        \\ { \green{ // add back original sample mean } }
        \STATE $\X \gets \X + \bm{1}' \bmu$
    \end{algorithmic}
\end{algorithm}

\subsubsection{Convergence with unknown variance} 
Note that \eref{eq:lr_alpcah} is a nonconvex function and we apply alternating minimization,
also known as block coordinate descent or block nonlinear Gauss-Seidel method,
to solve the optimization problem. 
Given a noise variance lower bound $\epsilon > 0$,
the feasible sets for $\L,\R,\Pi$ variables are given by
\begin{align}
    \mathcal{S}_{L} & = \mathbb{R}^{D \times \hat{d}}, \quad
    \mathcal{S}_{R} = \mathbb{R}^{N \times \hat{d}} \\
    \mathcal{S}_{\Pi} &= \{ \Pi_{i,j} \in [\epsilon,\infty) \hspace{2mm} \forall i=j, \hspace{1mm} 0 \text{ o.w.} \}. 
\end{align}
Given the following optimization problem with 
\begin{align}
    &\arg \min f(\L, \R, \Pi)     \label{eq:gs_problem}\\ 
    &\text{subject to } \L, \R, \Pi \in \mathcal{S} = \mathcal{S}_{\L} \times \mathcal{S}_{\R} \times \mathcal{S}_{\Pi}, \nonumber
\end{align}
the following theorem establishes
local convergence of $\{(\L_t,\R_t,\Pi_t)\}_{t=1}^T$
to critical points of \eref{eq:gs_problem}.

\begin{theorem}
\label{thm:lr-alpcah}
The sequence generated by alternating minimization 
$\{(\L_t,\R_t,\Pi_t)\}_{t=1}^T$ in \aref{alg:lr-alpcah}
has limit points that are critical points of \eref{eq:gs_problem}.
\end{theorem}
\begin{proof}
The cost function $f$ is a continuously differentiable function by inspection of both terms.
The feasible sets $\mathcal{S}_L,\mathcal{S}_R$ are trivially nonempty, closed, and convex sets by definition.
Moreover, $\Pi \in \mathcal{S}_{\Pi}$ since it is an enforced constraint of the optimization in \eref{eq:gs_problem}. 
The function $f$ is componentwise strictly quasiconvex with respect to the two blocks \L and \R.
This is because $f(\L,\R,\Pi) \text{ w.r.t. } \L$ and $f(\L,\R,\Pi) \text{ w.r.t. } \R$ are convex terms
it follows that they are pseudoconvex functions \cite{pseudoconvex}
and this implies they are also strict quasiconvex 
functions \cite{pseudoconvex}.
It then follows from \cite[Prop.~5]{grippo2000convergence}
that the sequence generated by alternating minimization
$\{(\L_t,\R_t,\Pi_t)\}_{t=1}^T$
converges to limit points that are also critical points of \eref{eq:gs_problem}.
\end{proof}

\begin{figure*}
\centering
\subcaptionbox{Ratio of subspace affinity errors LR-ALPCAH/PCA
(known variance, no cross-validation required)
\label{fig:pca}}
{
\includegraphics[width=0.45\columnwidth]{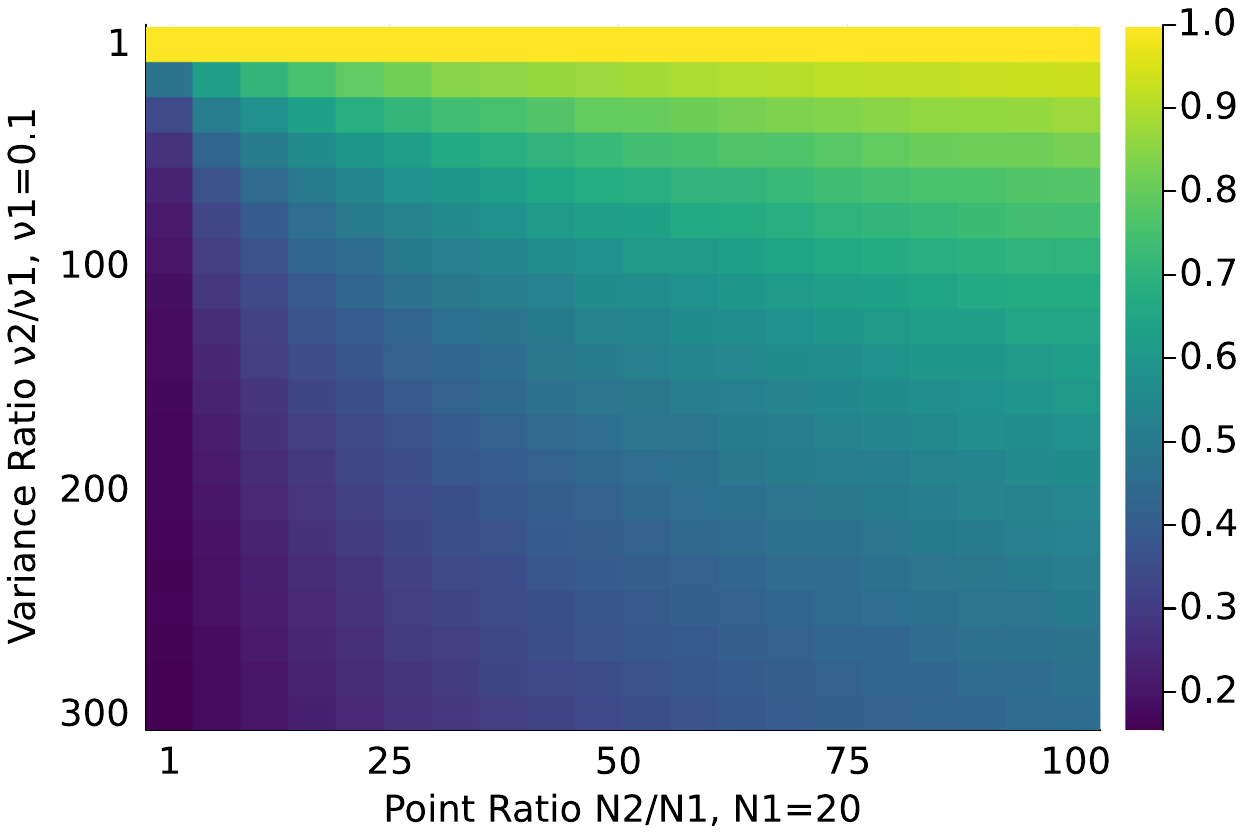}
}
\hfil
\subcaptionbox{Ratio of subspace affinity errors LR-ALPCAH/PCA-GOOD
(PCA using good data only and LR-ALPCAH using all of the data)
\label{fig:pca_good}}
{
\includegraphics[width=0.45\columnwidth]{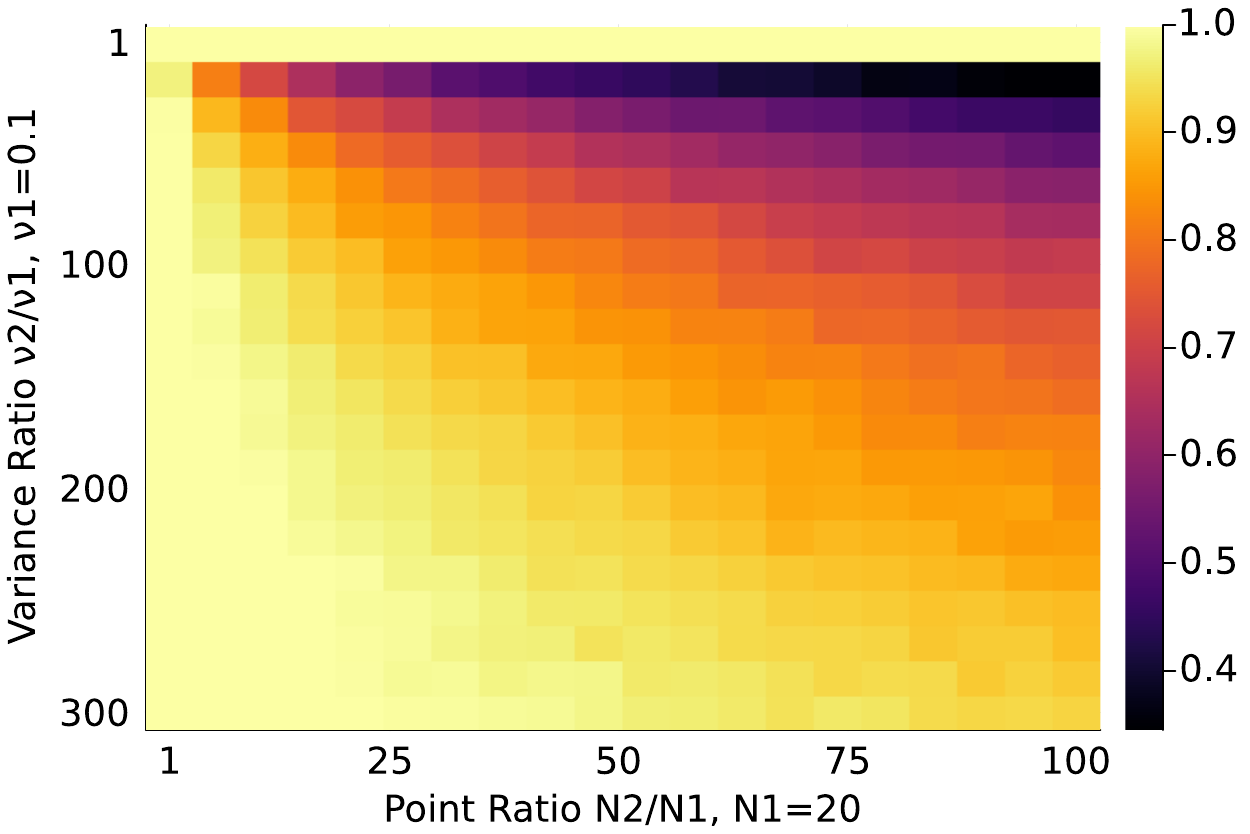}
}

\caption{Subspace affinity error
$ \normfrobr{ \U \U' - \hU \hU'} / \normfrobr{ \U \U' } $
performance of LR-ALPCAH compared to PCA.}
\label{fig:relative_heatmaps}
\end{figure*}

\begin{figure}
\centering
\includegraphics[width=0.95\columnwidth]{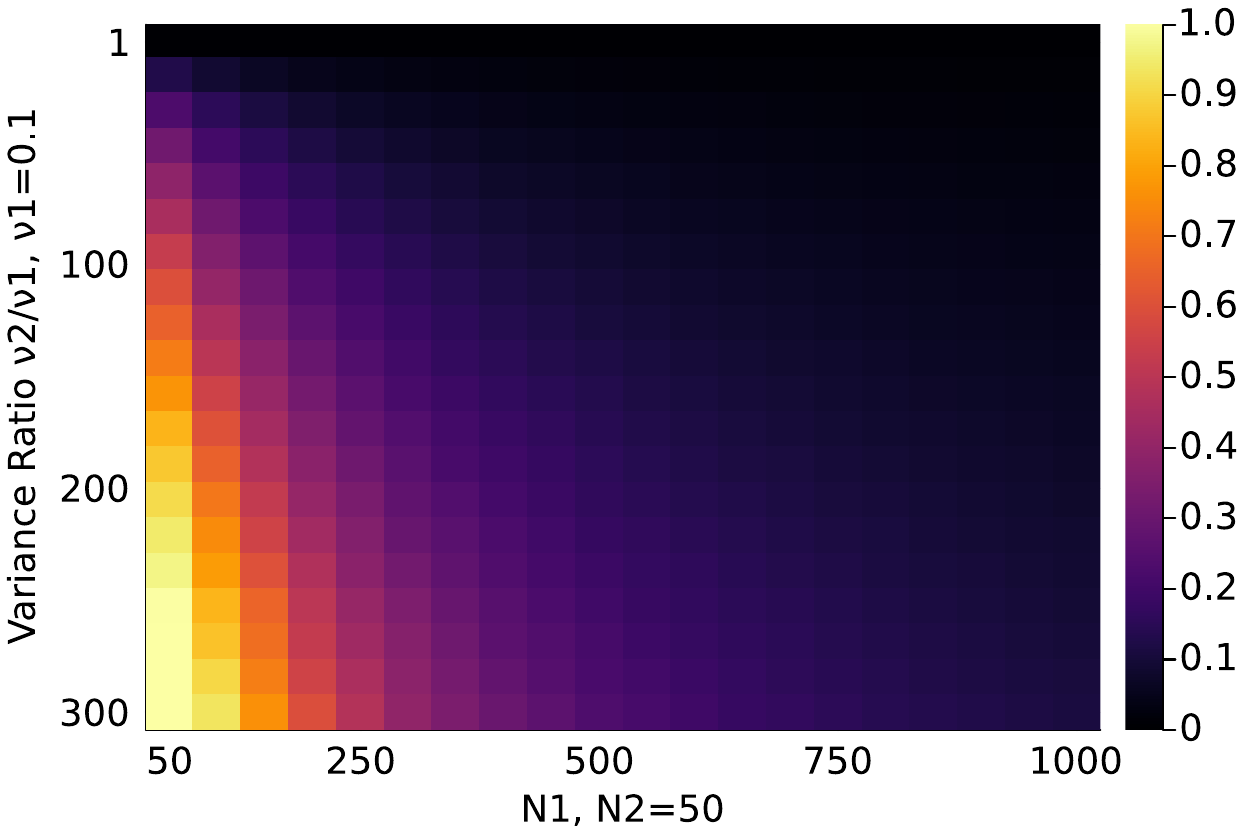}
\caption{Absolute difference of LR-ALPCAH subspace error
subtracted from PCA while good data amount varies.}
\label{fig:good_data_alpcah}
\end{figure}

\section{Empirical Results \& Discussion}
\label{paper:experiments}

This section summarizes synthetic and real data experiments,
including astronomy spectra and RNA sequencing data,
that explore various aspects of subspace learning
from heteroscedastic data.

\subsection{Synthetic Experiments}

\begin{figure}[t]
    \centering
    \includegraphics[width=0.95\linewidth]{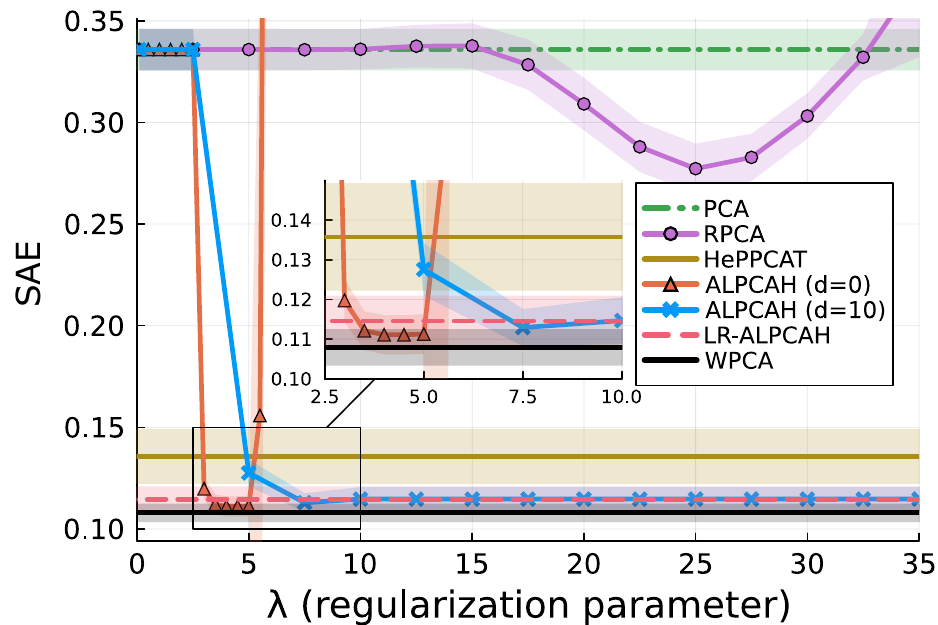}
    \caption{Absolute subspace quality performance of ALPCAH compared against other methods. Zoomed-in areas shown within plots for better visibility for certain $\lambda$ ranges.}
    \label{fig:absolute_error_unknown}
\end{figure}

\begin{figure*}
\centering
\subcaptionbox{Subspace learning results showing SAE
for ALPCAH and LR-ALPCAH relative to other methods.
\label{fig:astro_subspace}}
{
\includegraphics[width=0.5\columnwidth]{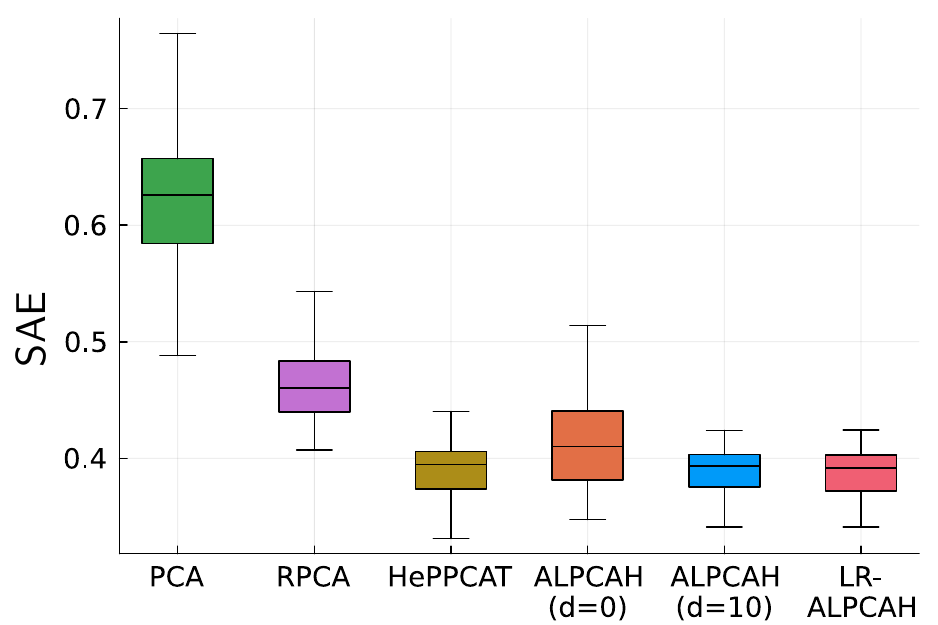}
}
\hfil
\subcaptionbox{Noise standard deviations estimates of LR-ALPCAH
compared to known astronomical reference values.
\label{fig:astro_variances}}
{
\includegraphics[width=0.45\columnwidth]{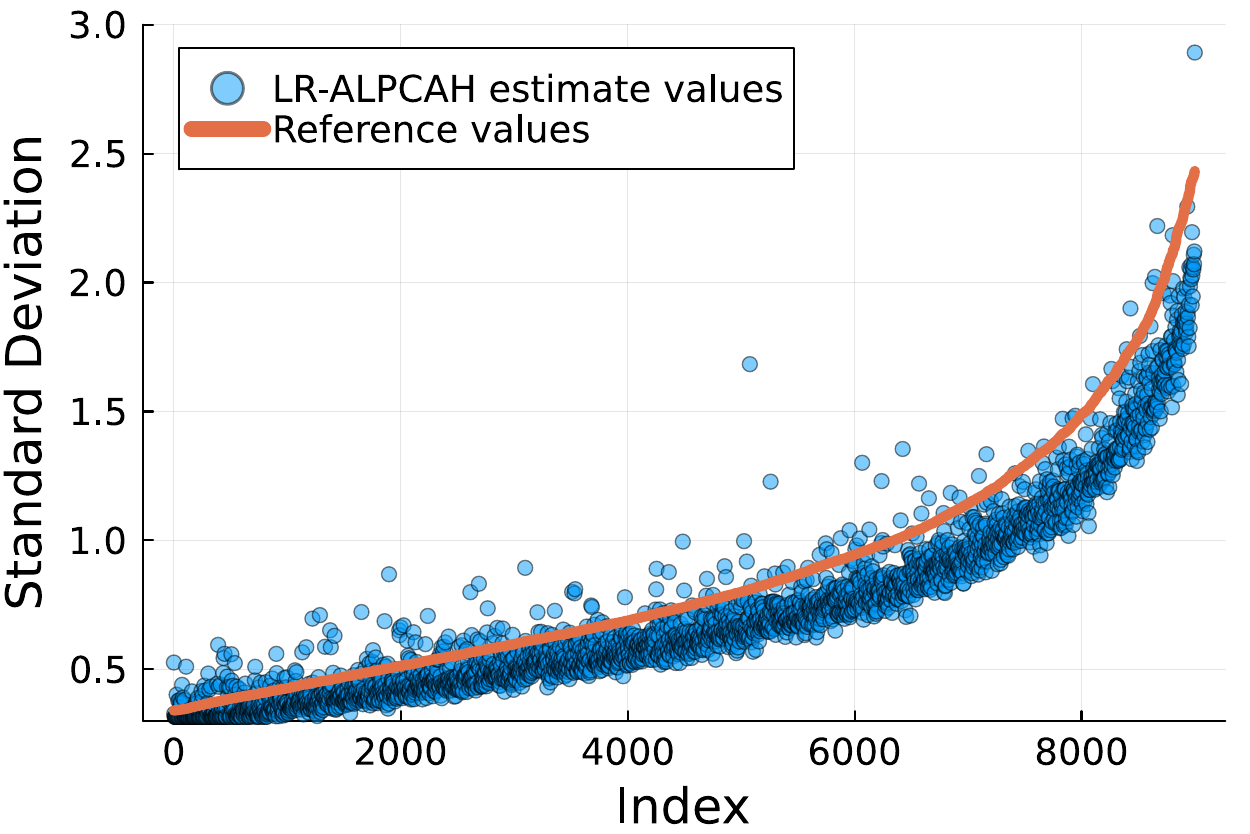}
}

\caption{Experimental results of quasar flux data
for subspace learning and noise sample estimation.}
\label{fig:astro_pca_results}
\end{figure*}

\begin{figure}
\centering
\includegraphics[width=0.8\textwidth]{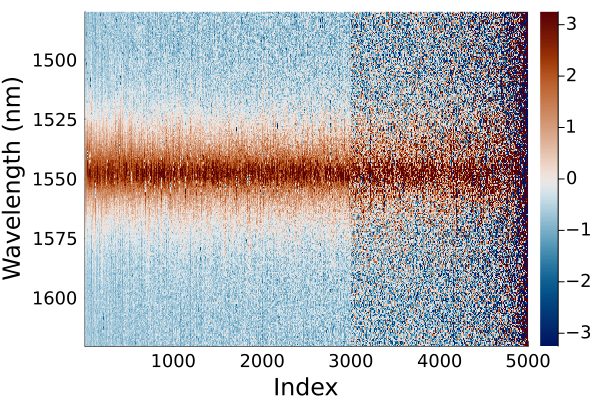}
\caption{Sample data matrix of quasar flux measurements across wavelengths
for each (column-wise) sample.}
\label{fig:astro_spectra}
\end{figure} 

This section uses synthetic data
to compare LR-ALPCAH with other methods.
We begin by describing the experimental setup,
followed by an investigation of PCA,
and after that compare to RPCA, HePPCAT, and WPCA.

\paragraph{Experimental Setup} 
We consider two groups of data, one with fixed quality,
meaning fixed size and additive noise variance, and one whose parameters we vary.
Let $\yi \in \mathbb{R}^{100} $ be $D = 100$ dimensional ambient-space data.
Let $\U \in \mathbb{R}^{100 \times 5}$
denote a basis for a $d=5$ dimensional subspace
generated by random uniform matrices such that
$\U \Sig \V' = \mathrm{svd}(\A)$,
where $A_{ij} \sim \mathcal{U}[0,1]$.
We use the compact SVD here.
The low-rank data is simulated as $\x_i = \U \z_i$
where the coordinates $\z_i \in \mathbb{R}^{5}$
were generated from
$\mathcal{U}[-100,100]$ for each element.
Then, we generated
$\y_i = \U \z_i + \bepsilon_i$
where $\bepsilon_i \in \mathbb{R}^{100}$
was drawn from $\mathcal{N}(\0, \nu_i \I)$.
The error metric used is subspace affinity error (SAE)
that compares the difference in projection matrices
\begin{equation} 
\text{SAE}(\U,\hU) = \normfrobr{ \U \U' - \hU \hU' } / \normfrobr{ \U \U' }
\end{equation}
so that a low error signifies a closer estimate of the true subspace.
This metric is also known as normalized chordal distance \cite{chordaldistance}.
In summary, the noisy data $\Y = [\y_1,\ldots,\y_N]$ is generated accordingly,
an estimate $\Hat{\X}$ is generated from \eqref{eq:heteroscedastic_subspace_model},
the subspace basis is calculated by
$\Hat{\X} = \sum_i \Hat{\sigma}_i \Hat{\u}_i \Hat{\v_i}'
\implies \hU = [\Hat{\u}_1, \ldots, \Hat{\u}_d]$,
and we report the subspace affinity error.

\paragraph{Subspace Basis Estimation (LR-ALPCAH vs. PCA)} 
We explored the effects of data quality and data quantity
on the heteroscedastic subspace basis estimates
in different situations.
For the heatmaps in \fref{fig:relative_heatmaps},
we focused on comparing LR-ALPCAH with PCA only
to discuss this method in the general context of subspace learning.
In \fref{fig:relative_heatmaps},
each pixel represents the ratio
$\text{SAE}(\U,\hU_{\text{LR-ALPCAH}}) / \text{SAE}(\U,\hU_{\text{PCA}})$.
A value close to 1 implies LR-ALPCAH did not perform much better than the other method,
whereas a ratio closer to 0 implies LR-ALPCAH performed relatively well.
The average SAE ratio of 50 trials is used,
where each trial has different noise, basis coefficients, and subspace basis realizations.
The noise variance for group 1 is fixed to $\nu_1 = 0.1$ with $N_1 = 20$ point samples.
We varied group 2 point samples $N_2$ and noise variances $\nu_2$,
as illustrated in the x-axis and y-axis respectively for the heatmaps shown.

\fref{fig:pca} compares LR-ALPCAH against PCA
in the situation where noise variances are known.
In this case, LR-ALPCAH performs well relative to PCA in noisy situations
and can improve estimation, especially in extreme heteroscedastic regions.
From the bottom left corner and moving rightwards,
the estimation error worsened
as the number of noisy points increased.
To clarify, LR-ALPCAH never performed worse than PCA,
only that the advantage gap decreased as more noisy samples were added.
This means that, in LR-ALPCAH, the noisy points may have contributed too much to the estimation process when the good quality data should have more influence in the process.
For these results, we used the inverse noise variances as the weighing scheme
as this is a natural choice that arises from the Gaussian likelihood.
However, finding the optimal scheme to mitigate this worsening effect is a topic of future work.

\begin{figure*}
\centering
\includegraphics[width=0.98\textwidth]{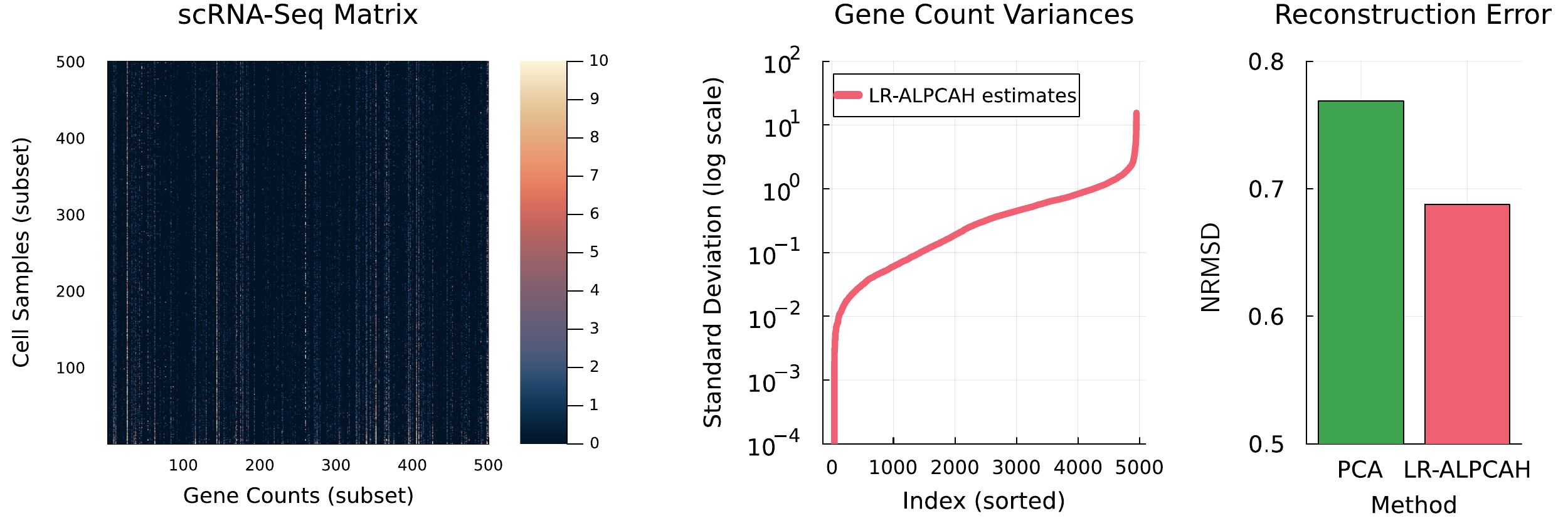}
\caption{
Biological scRNA-seq data results.
}
\label{fig:bio_combined}
\end{figure*} 




\fref{fig:pca_good} is similar to \fref{fig:pca}
but only using the high quality points for PCA specifically,
whereas LR-ALPCAH used all of the data.
One can see that even when there was enough good data,
there was still an improvement relative to applying PCA to just the good data alone.
The improvement increased as more noisy points were added.
Thus, it is beneficial to collect and use all of the data,
since the noisy points offer meaningful information
that can improve the estimate of the basis
versus using good data alone,
especially in data-constrained situations. 

\paragraph{Effects of Good Data}
\fref{fig:good_data_alpcah} explores how the number of good data samples
affects subspace learning quality.
We fixed $N_2 = 50$ and varied $N_1$ while keeping $\nu_1 =0.1$ and varying $\nu_2$.
This figure plots the difference
$\text{SAE}(\U,\hU_{\text{PCA}}) - \text{SAE}(\U,\hU_{\text{LR-ALPCAH}})$
to see when it is advantageous to use LR-ALPCAH instead of PCA.
In the absolute sense, both methods performed similarly when good data is abundant.
However, when good data was more limited,
there were larger differences in subspace quality,
meaning it is more advantageous to use LR-ALPCAH.

\paragraph{Absolute Subspace Error}
In this section, we discuss absolute error of the algorithms in the unknown noise variance setting without group knowledge. For \fref{fig:absolute_error_unknown}, we fixed $N_1 = 50, N_2 = 450$ that have noise variances $\nu_1 = 0.25, \nu_2 = 100$.
The regularization parameter $\lambda$ is varied (ALPCAH \& RPCA only) and subspace dimension is $d=10$.
As before, we use the subspace affinity error $\text{SAE}(\U,\Hat{\U})$.
The average error is plotted out of 50 trials with standard deviation bounds
for each $\lambda$ value. \fref{fig:absolute_error_unknown} represents the unknown variance case but we again use WPCA with weights $w_i = \nu_i^{-1}$,
a known variance method,
to illustrate the lowest possible affinity error if one hypothetically knew the noise variances. 

In \fref{fig:absolute_error_unknown},
when using rank knowledge, ALPCAH ($\hat{d}=10$) approaches the error of the other methods
as $\lambda$ grows. When not using rank knowledge, for ALPCAH ($\hat{d}=0$), the method can perform just as well but requires cross-validation to find an ideal $\lambda$ range.
Both ALPCAH ($\hat{d}=10$) and LR-ALPCAH achieved lower error than HePPCAT, likely because there are no distributional assumptions on the basis coefficients with ALPCAH/LR-ALPCAH. 
The RPCA method did not perform well in these experiments, likely because of the model mismatch between the heteroscedastic data and the outlier assumption of RPCA. 
Excluding the case when rank knowledge is not known, ALPCAH ($\hat{d}=0$), the regularization parameter appears to be robust to this landscape of different variance and point ratios.


\subsection{Real Data Experiments}
\label{sec:read-data1}

\subsubsection{Astronomy spectra data}
We investigated quasar spectra data
from the Sloan Digital Sky Survey (SDSS) Data Release 16 \cite{sdss}
using its DR16Q quasar catalog \cite{sdss2}.
Each quasar has a vector of flux measurements across wavelengths
that describes the intensity of observing that particular wavelength.
In this dataset, the noise is heteroscedastic across the sample space (quasars) and feature space (wavelength),
but we focused on a subset of data that is homoscedastic across wavelengths
and heteroscedastic across quasars.
The noise for each quasar is known
given the measurement devices used for data collection \cite{sdss},
but we performed estimation as if the variances were unknown
so that we could compare the estimated values to the reference values.
We preprocessed the data
(filtering, interpolation, centering, and normalization)
based on supplementary material 5 of \cite{optimal_pca}.
We formed a training dataset based on the 1000 smallest variance quasar flux samples
and performed PCA to get a ``ground-truth'' measurement of the subspace basis
using $\hat{d}=5$ as the rank parameter estimated from SignFlipPA \cite{signFlips}.
We formed the test dataset by excluding the 1000 samples used during training
and combining 9000 samples of various noise quality,
leading to heteroscedasticity across samples as shown in \fref{fig:astro_spectra}.
This figure shows only the 3000 lowest noise variance data samples
along with 2000 noisier samples to illustrate the differences in data quality.

\begin{table}[t]
\begin{tabular}{|l|r|r|r|}
\hline
\multicolumn{1}{|c|}{}
& \multicolumn{1}{c|}{\begin{tabular}[c]{@{}c@{}}Time\\ (ms)\end{tabular}}
& \multicolumn{1}{c|}{\begin{tabular}[c]{@{}c@{}}Memory\\ (MiB)\end{tabular}}
& \multicolumn{1}{c|}{\begin{tabular}[c]{@{}c@{}}Mean\\ SAE\end{tabular}} \\
\hline
\multicolumn{4}{|c|}{Homoscedastic Reference}
\\ \hline
PCA & 32.4 & 7.3 & 0.65
\\ \hline
\multicolumn{4}{|c|}{Heteroscedastic Methods}
\\ \hline
RPCA (classical) & 5091.8 & 7977.5 & 0.46 \\ \hline
HePPCAT & 1339.1 & 5731.6 & 0.39 \\ \hline
ALPCAH ($\hat{d}$=0)   & 4339.3 & 3838.6 & 0.41 \\ \hline
ALPCAH ($\hat{d}$=10) & 4339.9 & 3838.8 & 0.39 \\ \hline
LR-ALPCAH   & \textbf{153.5} & \textbf{459.0} & \textbf{0.38} \\ \hline
\end{tabular}
\caption{Subspace learning results on quasar flux data.}
\label{fig:tab_astro_pca}
\end{table}

For our subspace quality experiment,
we report SAE using the ``ground-truth'' basis
over 100 trials for various methods.
We ran each applicable subspace learning algorithm for 100 iterations to ensure convergence.
In \fref{fig:astro_subspace},
RPCA seems to perform slightly worse than the other methods,
indicating a model mismatch between outliers and heteroscedastic data.
Moreover, it seems that LR-ALPCAH and ALPCAH performed equally well as HePPCAT
in this specific real data example.
All methods performed better than PCA,
indicating a mismatch between the homoscedastic assumption of PCA
and the heteroscedastic data.
Additionally, we examined the computational time and memory requirements
for these methods on this test dataset.
\tabref{fig:tab_astro_pca} shows that the proposed LR-ALPCAH method
is both extremely fast and memory efficient
relative to the other heteroscedastic methods as shown in bold.
Since we have reference noise variance values,
we also examined how the estimated noise variance values compared to the reference values.
\fref{fig:astro_variances} sorts the data based on the reference variance values
and plots the ALPCAH estimates.
ALPCAH estimates generally tracked the global trend found in the reference values
but there are minor variations among adjacent points.

\subsubsection{Biological scRNA-seq data}
\label{sec:rna_seq}
This section applies PCA and LR-ALPCAH to real data
from single-cell RNA-sequencing data (scRNA-seq)
from 
\cite{rnaseq}.
This sequencing technology is useful
for quantifying the transcriptome of individual cells \cite{rnaseq2}.
The data is high dimensional
since thousands of genes are counted for thousands of cell samples,
which produces challenges for data analysis.
PCA methods are useful for scRNA-seq data
to perform gene variation analysis and clustering
in low-dimensional spaces to study gene groups \cite{rnaseq3}.
Heterogeneous noise may occur among both cells and genes \cite{signFlips},
which prompts further investigation into heteroscedastic-aware PCA methods on scRNA-seq data.
The data matrix consists of 10,000 cells by 5,000 genes.
We preprocessed the data by subtracting the mean
and replacing the missing values with zeros.
Since the noise variances are unknown in this application,
we cannot have a ``ground truth" subspace to compare against.
Instead, we separate the data into train and test, and calculate the NRMSD to compare reconstruction quality, i.e.,
\begin{equation}
    \text{NRMSD} = \normfrobr{ \Y_{\text{test}} - \U_{\text{train}} \U_{\text{train}}' \Y_{\text{test}} }
    \hspace{1mm} / \hspace{1mm} \normfrobr{ \Y_{\text{test}} } .
\end{equation}
The subspace basis was learned on the training data with PCA or LR-ALPCAH
and the test data was used to assess reconstruction quality
by projecting test data onto the subspace basis
and using the basis coefficients to return to the ambient space.
In this experiment, SignFlipPA was used to determine an appropriate rank
\cite{signFlips}.
\fref{fig:bio_combined} shows a subset of the data matrix.
Here, the color map is clipped to 10 to better visualize the matrix as most gene counts are sparse.
The middle plot shows sorted noise variances estimated by LR-ALPCAH
indicating some potential heterogenoity by one or two orders of magnitude.
The right plot shows that LR-ALPCAH has a better reconstruction quality
since it has $\sim$0.1 lower NRMSD than PCA.
The difference between PCA and LR-ALPCAH is more modest with this dataset.
Possibly, the results could be improved further
by developing a method that handles heteroscedasticity across both the samples and features,
as this data is doubly heteroscedastic.
Moreover, real scRNA-seq data has additional challenges
such as dependent noise that is not modeled by our method.
However, preliminary results indicate that LR-ALPCAH is a promising approach
and further investigation into addressing model assumptions
is
an interesting direction of future work.
\section{Conclusion}
\label{paper:conclusion}
This paper proposed two subspace learning algorithms that are robust to heteroscedasticity
by jointly learning the noise variances and subspace bases.
While
LR-ALPCAH is memory efficient and fast,
its application is limited to sample-wise heteroscedasticity.
It would be interesting to generalize this work to be doubly heteroscedastic, 
where
the features themselves also have different noise variances.
Applications such as biological sequencing \cite{hetero_feature1}
and photon imaging \cite{hetero_feature2} could benefit from such an extension. In the scRNA-seq application, we computed missing entries as zeros, which is a natural choice for low-rank models. However, others have worked on adapting PCA methods for missing data \cite{gilman_missingdata} so such an approach could be beneficial given the higher than expected NRMSD with LR-ALPCAH. 
This generalization is nontrivial
so it is left for future work.
Additionally, our model and the comparison methods
are limited to the subspace setting,
but some applications
like resting-state functional MRI \cite{functional_mri_vae}
benefit from manifold learning approaches
\cite{functional_mri_vae}.
It would be interesting to explore other approaches
such as a heteroscedastic variational autoencoder \cite{betavae}
to expand the possible applications of heteroscedastic data learning.

\section{Acknowledgments}
This work is supported in part by NSF CAREER Grant CCF-1845076 and NSF Grant CCF-2331590.
There are no financial conflicts of interest in this work.
We thank Jason Hu for helpful discussions related to algorithm convergence for \tref{thm:alpcah}.
We thank David Hong
for suggesting heteroscedastic data applications
such as astronomy spectra and single-cell RNA sequencing.

\section{Appendix}
\subsection{PCA Bound Experiment}

\begin{figure}[H]
\centering
\includegraphics[width=0.98\textwidth]{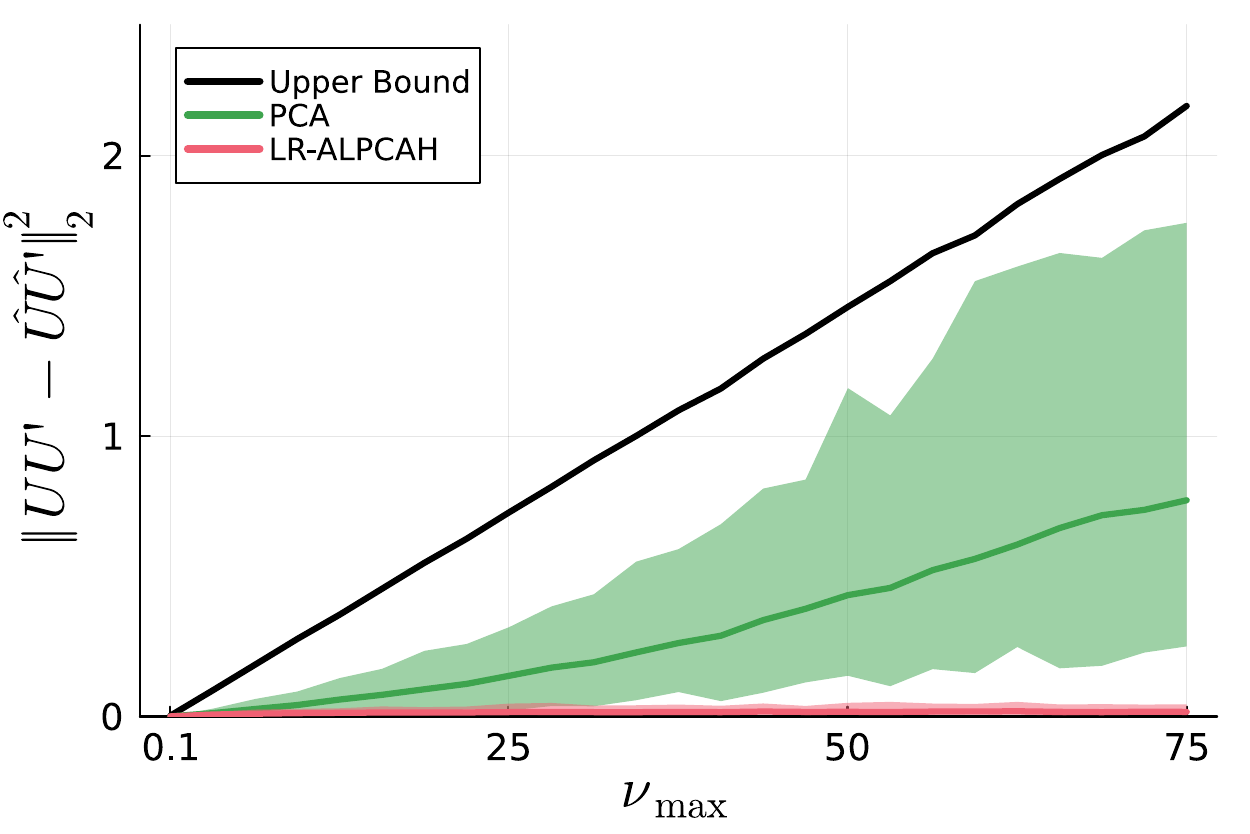}
\caption{
Experimental verification of heteroscedastic impact on PCA upper bound \eref{thm:pca_bound}.
}
\label{fig:bounds}
\end{figure}

This section focuses on providing empirical verification
of our subspace bound in \eref{thm:pca_bound}.
Before doing so, we mention that the random matrix theory bound in \eref{eq:random_matrix_bound}
depends on a universal constant $c_1$,
independent of $D$ and $N$,
that is not calculated in the source paper \cite{normbound}.
Let $A = [a]$ be a $1 \times 1$ matrix with element $a \sim \mathcal{N}(0,1)$.
For the LHS in \eref{eq:random_matrix_bound},
the spectral norm in this instance is $\|A\|_2 = |a|$.
This implies that
$\mathbb{E} [\|A\|_2]$
is equivalent to calculating the mean of a folded normal distribution.
Since $a$ is a standard normal random variable,
$\mathbb{E} [|a|] = \sqrt{\Frac{2}{\pi}}$.
One can verify
from
\eref{eq:rmt_bound1} \eref{eq:rmt_bound2} \eref{eq:rmt_bound3}
that
the RHS in \eref{eq:random_matrix_bound}
simplifies to $2+\sqrt[4]{3}$.
Solving for $c_1$,
the inequality becomes
$c_1 \geq \Frac{\sqrt{2/\pi}}{(2+\sqrt[4]{3})} \approx 0.24$.
In our subspace bound \eref{thm:pca_bound},
both sides are squared
and a factor of $2$ exists in \eref{eq:daviskahan},
therefore the constant in \eref{thm:pca_bound} is
$c = 4c_1^2 \approx 0.22$.
Knowing this constant,
it is now possible to experimentally verify \eref{thm:pca_bound}.
The experimental setup consists of generating random rank-$3$ subspaces
within a $100$ dimensional ambient space.
The data samples consist of two groups,
one with $n_1 = 30 \text{ samples}, \nu_1=0.1 $
and the other with $n_2 = 970 \text{ samples}$
and a varying $\nu_2 \in \{ 0.1, \ldots,75 \}$.
During the course of $50$ trials,
we computed the mean spectral norm projection error,
i.e.,
$\|\hU \hU'-\U \U'\|_2^2$,
along with the minimum and maximum error values
for that $\nu_2$ instance.
\fref{fig:bounds}
illustrates
that PCA scales similarly to the bound in \eref{thm:pca_bound},
yet our method, LR-ALPCAH,
empirically did not degrade at the same rate,
indicating robustness to heterocedasticity.

\bibliographystyle{IEEEtran}
\bibliography{setup/ref.bib}

%

\begin{IEEEbiography}[{\includegraphics[width=1in,height=1.25in,clip,keepaspectratio]{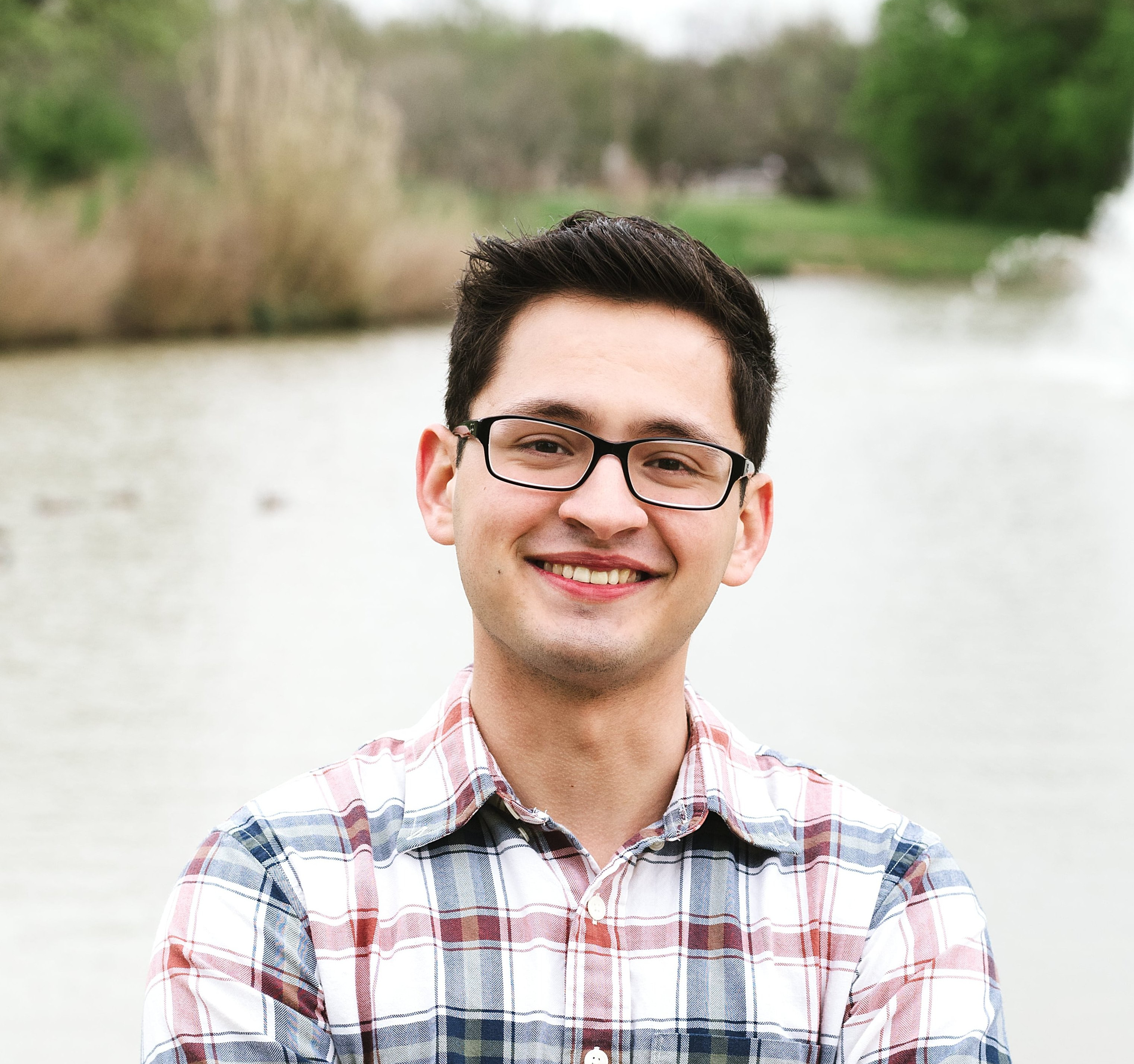}}]{Javier Salazar Cavazos}
(Student Member, IEEE) received dual
B.S. degrees in electrical engineering and mathematics 
from the University of Texas, Arlington, TX, USA, 2020 and the M.S. degree in electrical and computer engineering (ECE) from the University of Michigan, Ann Arbor, MI, USA, 2023. He is
currently working toward the Ph.D. degree in electrical 
and computer engineering (ECE) working with Professor
Laura Balzano and Jeffrey Fessler both with the University of Michigan, Ann
Arbor, MI, USA. 
His main research interests include signal and image processing, machine learning, deep learning, and optimization. Specifically, current work focuses on topics such as low-rank modeling, clustering, and Alzheimer's disease diagnosis and prediction from MRI data.
\end{IEEEbiography}

\begin{IEEEbiography}[{\includegraphics[width=1in,height=1.25in,clip,keepaspectratio]{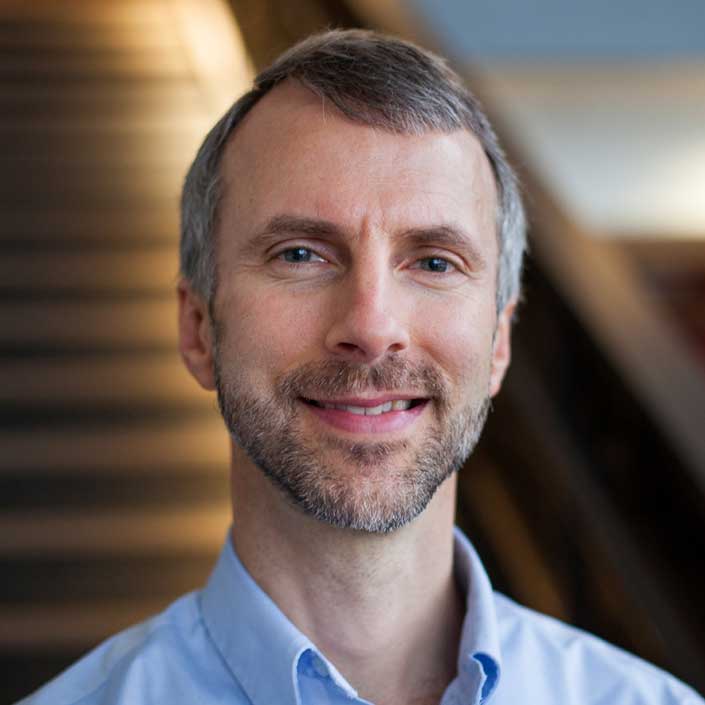}}]{Jeffrey A. Fessler}
(Fellow, IEEE) received the
B.S.E.E. degree from Purdue University, West
Lafayette, IN, USA, in 1985, the M.S.E.E. degree
from Stanford University, Stanford, CA, USA, in
1986, and the M.S. degree in statistics and the Ph.D.
degree in electrical engineering from Stanford University, in 1989 and 1990, respectively. From
1985 to 1988, he was a National Science Foundation
Graduate Fellow with Stanford. He is currently
the William L. Root Distinguished University Professor of EECS
and ECE Interim Chair
at the University of Michigan, Ann Arbor, MI, USA. 
His research focuses on various aspects of imaging
problems, and he has supervised doctoral research in PET, SPECT, X-ray CT,
MRI, and optical imaging problems. In 2006, he became a Fellow of the IEEE
for contributions to the theory and practice of image reconstruction. 
\end{IEEEbiography}


\begin{IEEEbiography}[{\includegraphics[width=1in,height=1.25in,clip,keepaspectratio]{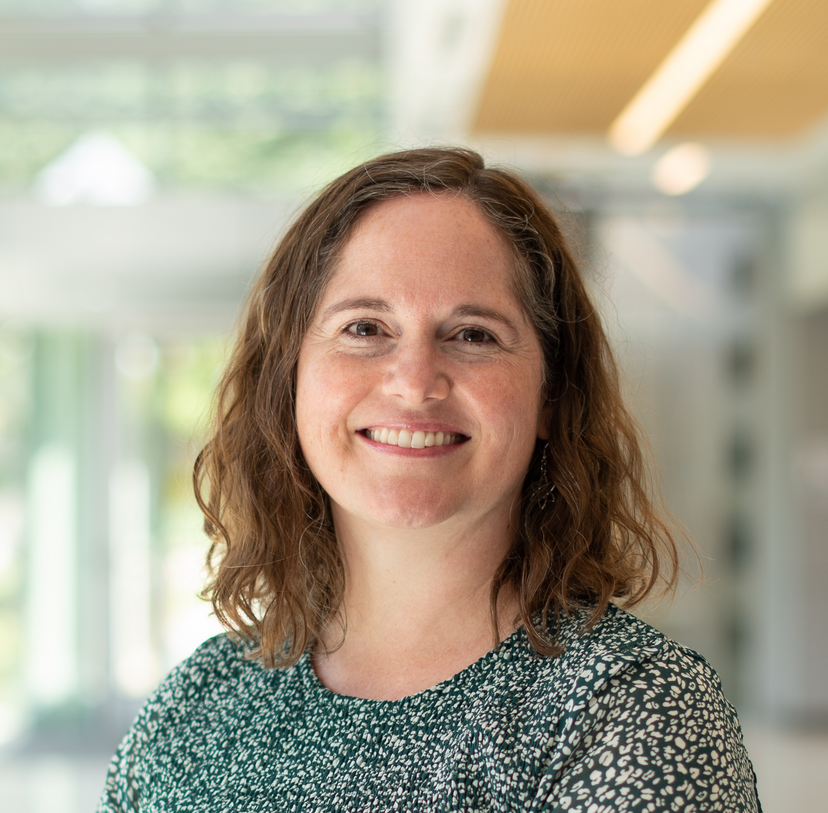}}]{Laura Balzano}
(Senior Member, IEEE) received the
Ph.D. degree in ECE from the University of Wisconsin, Madison, WI, USA. She is currently an Associate
Professor of electrical engineering and computer science with the University of Michigan, Ann Arbor, MI,
USA. Her main research focuses on modeling and optimization with big, messy data – highly incomplete or
corrupted data, uncalibrated data, and heterogeneous
data – and its applications in a wide range of scientific
problems. She is an Associate Editor for the
IEEE Open Journal of Signal Processing and
SIAM Journal of the Mathematics of Data Science. She was a recipient of
the NSF Career Award, the ARO Young Investigator Award, the AFOSR Young
Investigator Award. She was also a recipient of the Sarah Goddard Power Award at
the University of Michigan.
\end{IEEEbiography}





\end{document}